\documentclass[sigplan,nonacm]{acmart}

\usepackage{amsmath}
\usepackage{amsfonts}
\usepackage{booktabs}
\usepackage{graphicx}
\usepackage{multirow}
\usepackage{multicol}
\usepackage{tablefootnote}
\usepackage{float}
\usepackage{url}
\usepackage{makecell}
\usepackage{hyperref}
\usepackage[capitalize,noabbrev]{cleveref}
\AtBeginDocument{%
  }

\setcopyright{acmlicensed}
\copyrightyear{2018}
\acmYear{2018}
\acmDOI{XXXXXXX.XXXXXXX}
\acmConference[BIOKDD 2025]{24th International Workshop on Data Mining in Bioinformatics}{June 03--05,
  2018}{Woodstock, NY}
\acmISBN{978-1-4503-XXXX-X/2018/06}

\begin{document}

%%
%% The "title" command has an optional parameter,
%% allowing the author to define a "short title" to be used in page headers.
%\title{BioMedKG: Multimodal Contrastive Representation Learning in Augmented \underline{BioMed}ical \underline{K}nowledge \underline{G}raphs}

\title{Multimodal Contrastive Representation Learning in Augmented Biomedical Knowledge Graphs}

\author{Tien Dang}
\authornote{Both authors contributed equally to this work.}
\affiliation{
  \institution{University of Alabama at Birmingham}
  \city{Birmingham}
  \state{Alabama}
  \country{USA}
}

\author{Viet Thanh Duy Nguyen}
\authornotemark[1]
\affiliation{
  \institution{University of Alabama at Birmingham}
  \city{Birmingham}
  \state{Alabama}
  \country{USA}
}

\author{Minh Tuan Le}
\affiliation{
  \institution{Washington University in St. Louis}
  \city{St. Louis}
  \state{Missouri}
  \country{USA}
}

\author{Truong-Son Hy}
\authornote{Corresponding Author}
\affiliation{
  \institution{University of Alabama at Birmingham}
  \city{Birmingham}
  \state{Alabama}
  \country{USA}
}
\email{thy@uab.edu}

%%
%% By default, the full list of authors will be used in the page
%% headers. Often, this list is too long, and will overlap
%% other information printed in the page headers. This command allows
%% the author to define a more concise list
%% of authors' names for this purpose.
\renewcommand{\shortauthors}{Dang and Nguyen et al.}

%%
%% The abstract is a short summary of the work to be presented in the
%% article.
\begin{abstract}
Biomedical Knowledge Graphs (BKGs) integrate diverse datasets to elucidate complex relationships within the biomedical field. Effective link prediction on these graphs can uncover valuable connections, such as potential novel drug-disease relations. We introduce a novel multimodal approach that unifies embeddings from specialized Language Models (LMs) with Graph Contrastive Learning (GCL) to enhance intra-entity relationships while employing a Knowledge Graph Embedding (KGE) model to capture inter-entity relationships for effective link prediction. To address limitations in existing BKGs, we present PrimeKG++, an enriched knowledge graph incorporating multimodal data, including biological sequences and textual descriptions for each entity type. By combining semantic and relational information in a unified representation, our approach demonstrates strong generalizability, enabling accurate link predictions even for unseen nodes. Experimental results on PrimeKG++ and the DrugBank drug-target interaction dataset demonstrate the effectiveness and robustness of our method across diverse biomedical datasets. Our source code, pre-trained models, and data are publicly available at \url{https://github.com/HySonLab/BioMedKG}.
\end{abstract}

%%
%% The code below is generated by the tool at http://dl.acm.org/ccs.cfm.
%% Please copy and paste the code instead of the example below.
%%


\maketitle

\section{Introduction} \label{sec:introduction}
\begin{figure}[t]
    \centering
    \includegraphics[width=0.8\columnwidth]{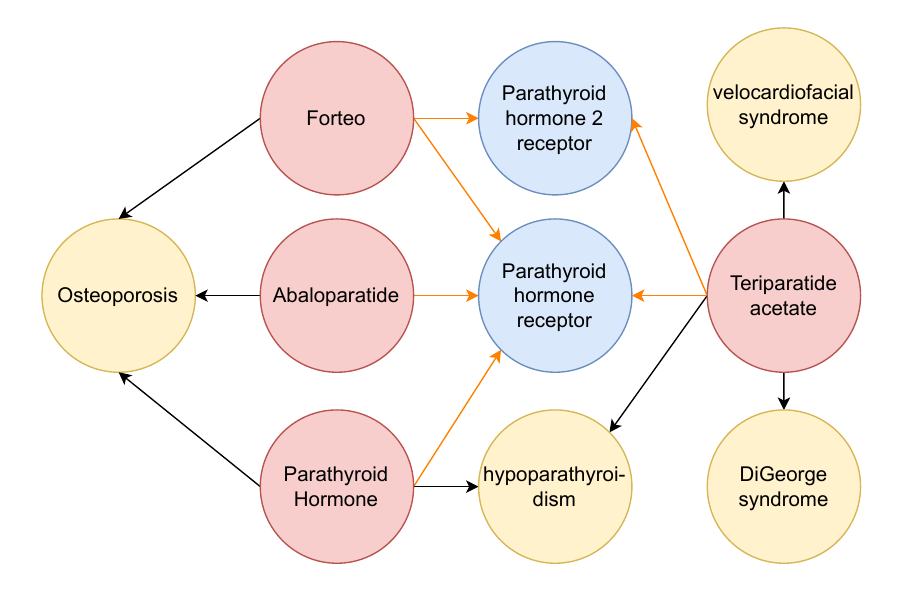}
       \caption{The subgraph illustrates the interactions surrounding the Parathyroid hormone receptor and its connections to related drugs and diseases. Different entity types are color-coded: red nodes represent drugs, blue nodes indicate genes or proteins, and yellow nodes denote diseases. Black arrows depict drug-treatment relationships with diseases, while orange arrows represent drug-receptor interactions. This subgraph is a focused segment of a broader Biomedical Knowledge Graph, which captures the complex interconnections among various biological entities.}
    \label{fig:BKG}
    \vspace{-0.6cm}
\end{figure}

BKGs are structured networks that represent intricate relationships among biological entities such as genes, proteins, diseases, and drugs (see \cref{fig:BKG}). Accurate link prediction within these graphs is crucial for identifying hidden relationships, discovering potential therapeutic targets, and suggesting drug repositioning opportunities \cite{BKGs_applications, biolp2, ngo2022predicting}. These capabilities can significantly accelerate biomedical research, leading to faster clinical advancements and more effective treatments.

Despite their potential, generating consistent and effective node representations for link prediction in BKGs remains a challenging task. A promising strategy to address this issue is enhancing the existing knowledge base by integrating rich, multimodal domain-specific data associated with these entities.

Recent advances show that pre-trained LMs can act as foundational knowledge bases, storing vast amounts of factual information \cite{LMs_knowledge_bases1, LMs_knowledge_bases2, LMs_knowledge_bases4, LMs_knowledge_bases3, LMs_knowledge_bases5}. When used as initial embeddings, LMs provide a strong foundation for downstream tasks by incorporating pre-existing knowledge from biomedical texts and databases \cite{LMs_survey}. These models offer rich semantic information that can enhance graph representation learning. However, prior works on BKGs \cite{BioBLP, Otter-Knowledge} have primarily focused on using single-modality node representations for each node type (e.g., amino acid sequences for proteins, SMILES strings for drugs, and textual descriptions for diseases), overlooking the potential to integrate multiple modalities for each node type. Moreover, while LM-derived embeddings serve as initial representations for knowledge graphs, they often lack graph topology, necessitating fine-tuning to effectively capture graph structure.

In this work, we propose a novel pre-trained node representation model designed to enhance link prediction performance in BKGs. Our comprehensive framework leverages the capabilities of LMs to generate robust entity representations, while seamlessly integrating multimodal information to enrich the contextual understanding of relationships within the graph. Specifically, we unify LM-derived embeddings for each entity and employ GCL to optimize intra-node relationships by enhancing mutual information within individual node types. Additionally, we utilize a KGE model to capture inter-node information between different biological entities. A key feature of our approach is its generalizability, as the node embeddings generated by our framework encapsulate both semantic information from LMs and relational information from GCL. This dual integration ensures that embeddings maintain a rich contextual understanding, allowing the framework to generate meaningful representations even for unseen nodes, thereby facilitating more accurate link prediction for novel entities.

However, our approach requires a BKG with well-defined node attributes, which are absent in most existing BKGs that lack comprehensive attributes for each entity type \cite{PrimeKG, Biokg}. To address this limitation, we introduce \textit{PrimeKG++}, an enriched knowledge graph that builds upon PrimeKG \cite{PrimeKG}. PrimeKG++ enhances the original dataset by incorporating biological sequences for each entity type—amino acid sequences for proteins, nucleic acid sequences for genes, and SMILES strings for small molecules—along with comprehensive textual descriptions. This integration diversifies node attributes and improves the overall utility of the knowledge graph, providing a valuable public resource for future research in biomedical knowledge graphs.

It is important to clarify that the primary focus of this paper is not on achieving state-of-the-art results in downstream tasks such as link prediction. Instead, we aim to propose a pretrained node representation model and demonstrate its effectiveness through comprehensive experiments. To evaluate this, we employed existing models with and without our pretrained node representations as initial inputs. Our experiments show that our pretrained node representations lead to significant performance improvements compared to random initialization or Direct LM-derived embeddings. By leveraging SOTA models for link prediction, we ensured that our comparisons were rigorous and meaningful, demonstrating the added value of our pretrained node representations within an established and high-performing framework.

The contributions of this work are summarized as follows:
\begin{itemize}
\item We propose a comprehensive framework that leverages LMs and GCL to create robust, multimodal node embeddings for BKGs.
\item We present PrimeKG++, an augmented biomedical knowledge graph enriched with biological sequences and textual descriptions, offering a comprehensive resource for our work and the biomedical research community.
\item We validate the effectiveness and generalizability of our approach through extensive empirical results.
\end{itemize}

\section{Related works} \label{sec:related}

\subsection{Knowledge Graph Embedding}
In the field of BKGs, link prediction research aims to uncover connections among biological entities by analyzing their existing links and attributes \cite{biolp1, biolp2, biolp3, biolp4, biolp5}. Knowledge graph embeddings, representing entities and relations as vectors, have gained popularity for this task. While traditional models, such as ComplEx \cite{ComplEx} and RotatE \cite{Rotate} have shown promising results in this link prediction task, two key constraints hinder them: firstly, they focus solely on the graph structure, ignoring valuable entity attribute information; and secondly, their reliance on predetermined embeddings for mapping entities and relations in the lookup table complicates integration with new entities. These constraints motivate us to construct a heterogeneous biomedical knowledge graph with multimodal metadata.

\subsection{Biomedical Language Model}
In BKGs, entities can possess different modalities, such as text or biological sequences. Essentially, a molecular sequence is the exact order of smaller units (monomers) that make up a large molecule (biopolymer). Similar to a textual description, it inherently possesses a sequential relationship that LMs can effectively process. Recent methods rely on pre-trained language models such as BERT \cite{bert} as the backbone for the attribute encoder. Protein sequences, which are strings of amino acid letters, can be effectively processed by models like ESM-2 \cite{ESM-2} and ProteinBERT \cite{ProteinBERT}. For genes, which are represented by nucleotide sequences, specific language models such as Nucleotide Transformers \cite{Dalla-Torre2023.01.11.523679} and DNABERT \cite{DNABERT} are required. Chemical structures are often represented using SMILES strings, a linear text format, which can be interpreted by models like BARTSmiles \cite{BARTSmiles} and MoLFormer \cite{MolFormer}. For textual descriptions in the biomedical domain, models such as BioGPT \cite{BioGPT} and BioBERT \cite{BioBERT} are leveraged to extract high semantic meaning, providing improved understanding and analysis of biomedical text. These findings inspire us to explore the potential of LMs to extract semantic information into node features in BKGs.

\subsection{Graph Contrastive Learning}
Many Graph Neural Networks rely on supervised learning with labeled data, which is costly and labor-intensive. To address this, a few studies(e.g., DGI \cite{DGI}, MVGRL \cite{MVGRL}, GMI \cite{GMI}, and GRACE \cite{GRACE}) use contrastive learning techniques, introducing Graph Contrastive Learning for self-supervised graph representation learning. These methods aim to maximize mutual information between an anchor node and its semantically similar counterparts while minimizing it for dissimilar ones. In recent years, contrastive learning has gained traction in knowledge graph embedding. KGCL \cite{kgcl2022} integrates knowledge graph learning with user-item interaction modeling through a joint self-supervised learning approach, improving robustness and addressing data noise and sparsity in recommendation systems. KE-GCL \cite{KEGCL} incorporates contextual descriptions of entities and proposes adaptive sampling to refine the knowledge graph contrastive learning. MCLEA \cite{multimodal_gcl} unifies information from various modalities and uses contrastive learning for discriminative entity representations. However, multimodal contrastive learning has not yet been explored in BKGs. In this paper, we present a novel graph representation learning framework incorporating contrastive learning for biomedical knowledge graphs.
\section{PrimeKG++: An Augmented Knowledge Graph}
PrimeKG \cite{PrimeKG} is a multimodal knowledge graph tailored for precision medicine, encompassing over 100,000 nodes across various biological scales. It features more than 4 million relationships among these nodes, categorized into 29 distinct edge types. We chose PrimeKG for its enriched disease and drug nodes, which are augmented with clinical descriptors sourced from medical authorities. This enrichment supports our approach by providing a robust foundation for applying LM-derived embeddings, allowing for more precise and contextually relevant analyses in biomedical research. However, PrimeKG exhibits limitations, particularly in its lack of additional context or descriptive information for other biological entity types such as genes and proteins. This deficiency hinders the graph's capacity to fully represent the complex interactions and functions inherent in these biological components.

To address these limitations, we developed PrimeKG++, an enhanced version of PrimeKG that adds detailed information for three key node types: gene/protein, drug, and disease. PrimeKG++ categorizes drug data into two subtypes: molecules, represented with SMILES strings, and antibodies, identified by amino acid sequences. For the gene/protein node type, it includes protein-coding genes, annotated with amino acid sequences, and non-coding genes, represented with nucleotide sequences. Descriptions are collected for all subtypes of both drugs and genes/proteins, providing essential context. These enhancements are meticulously linked to authoritative sources such as Entrez Gene \cite{NCBI} at NCBI for genes/proteins and DrugBank \cite{DrugBank} for drugs, using identifiers from PrimeKG. This detailed information allows PrimeKG++ to analyze and distinguish these entities' complex interactions and functions with greater precision.
\section{Method} \label{sec:method}
% As illustrated in \cref{fig:framework_overview}, our proposed framework optimizes node representations within biomedical knowledge graphs through a systematic, multi-step process. Initially, we generate attribute embeddings using specialized Language Models (LMs) (see Section \ref{sec:specialized_LM}) tailored to the specific data modalities of the nodes. These diverse node-specific features are then integrated into a unified embedding space via the Attribute Fusion Projection (see Section \ref{sec:fusion_module}). Following this, the Contextual Relationship Projection (see Section \ref{sec:relationship_projection}) is employed to enhance and elucidate relationships within homogeneous biomedical subgraphs, focusing on node types such as drugs, diseases, or genes/proteins. Finally, these refined embeddings are utilized for link prediction tasks using our Knowledge Graph Embedding (KGE) model (see Section \ref{sec:model_architecture}), to improve the predictive accuracy and practical utility of the knowledge graph in biomedical research.
% \tien{propose}

\begin{figure*}[ht]
    \centering
    \includegraphics[width=\textwidth]{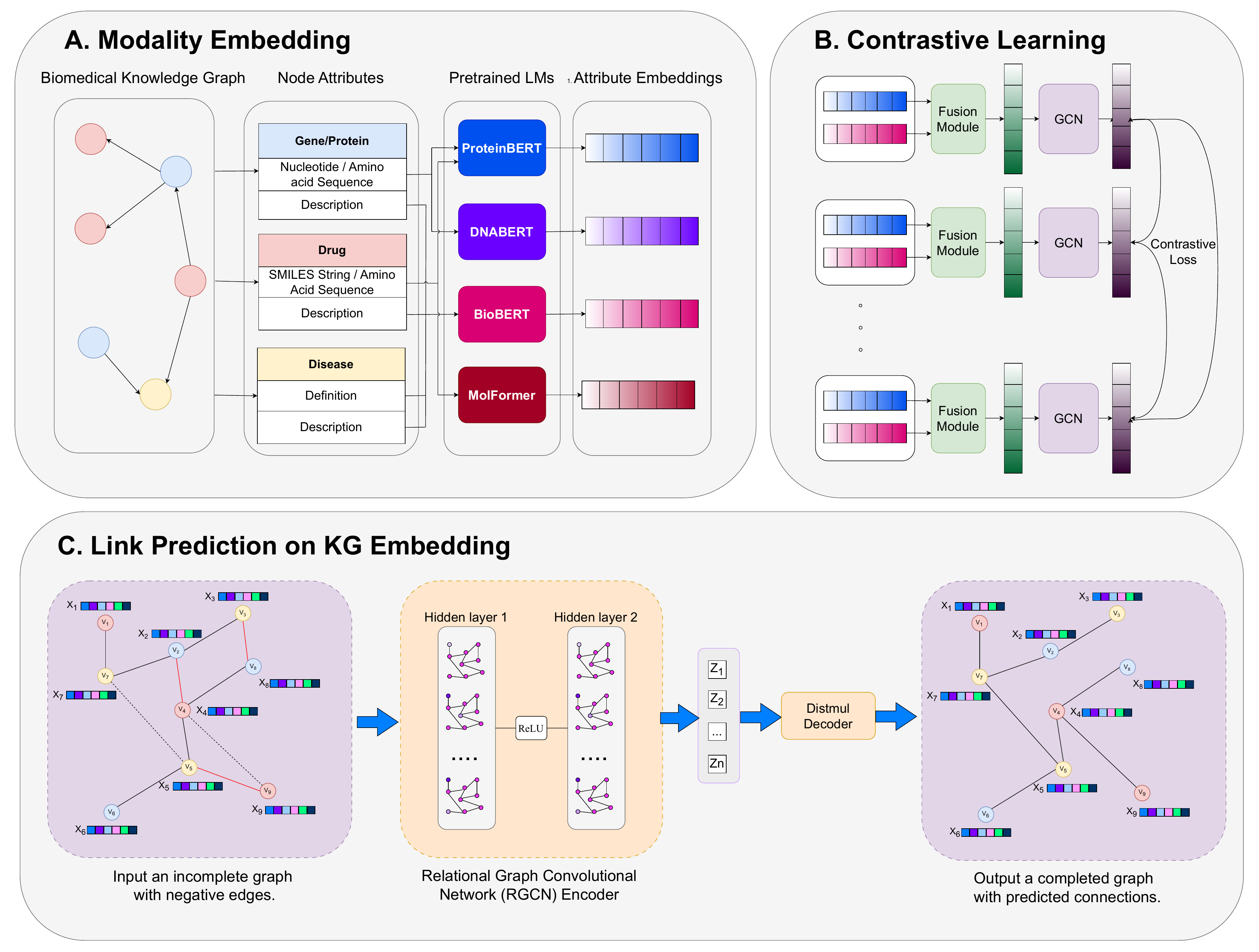}
       \caption{Overview of our proposed framework. \textbf{A. Modality Embedding}: Creating node attribute embeddings through domain-specific LMs. \textbf{B. Contrastive Learning:} Enhancement of LM-derived embeddings for specific node attributes of the same type through Fusion Module and Contrastive Learning. \textbf{C. Link Prediction on KG Embedding:} Utilizing the enhanced embeddings to perform link prediction tasks through a Knowledge Graph Embedding (KGE) model that learns relationships and enhances semantic information across distinct node types.}
    \label{fig:framework_overview}
    \vspace{-0.5cm}
\end{figure*}

Our framework is illustrated in \cref{fig:framework_overview}. Initially, we generate embeddings for each node type's modalities using their corresponding Language Models (Section \ref{sec:specialized_LM}). These modalities' embeddings are then integrated into a unified embedding space via the Fusion Module (Section \ref{sec:fusion_module}). Subsequently, the Graph Contrastive Learning module enhances relationships within homogeneous biomedical subgraphs, facilitating intra-node learning (Section \ref{sec:relationship_projection}). Finally, the Knowledge Graph Embedding module refines these embeddings through link prediction tasks to enhance learning across different node types, fostering inter-node learning (Section \ref{sec:model_architecture}).

\subsection{Preliminaries}
In the context of knowledge graphs where entities have associated attributes across various modalities, we define a Biomedical Knowledge Graph as \( G = (V, R, E, D, d) \). Here, \( V \) denotes the set of nodes, collectively \( \{v_1, \dots, v_n\} \), with n being the number of entities, \( R \) denotes the set of relations, and \( E \) consists of triples \( (h, r, t) \) where \( h, t \in V \) and \( r \in R \). The elements \( h \), \( r \), and \( t \) represent the head, relation, and tail of a triple, respectively. \( D \) represents a dataset with entities' attributes, where each entity type has specific attributes relevant to its biological role. The partial function \( d : V_d \rightarrow D \) maps a subset of entities \( V_d \subseteq V \), which have available attribute data, to their respective attributes, with \( d(v_i) \) retrieving the attribute data for an entity \( v_i \). This schema allows for tailored attribute representation, accommodating the diverse and specific data needs of different entity types within the graph.
% Too similar to BioBLP

\subsection{Modality Encoding} \label{sec:specialized_LM}

We utilize a set of \( k \) modality-specific \textit{encoders}, \( \{e_1, \dots, e_k\} \), where each encoder \( e_i \) corresponds to a pre-trained Language Model for a specific attribute modality \( D_i \subseteq D \). Each encoder \( e_i \) maps its respective attribute data into a distinct embedding space \( \mathcal{X}_i \), formally represented as \( e_i: D_i \rightarrow \mathcal{X}_i \). Selection of these LMs involves ensuring uniformity in embedding sizes and balancing the computational complexity with the desired level of accuracy, to optimize both integration across modalities and overall system efficiency. Specifically, we use ProtBERT \cite{ProteinBERT} for protein sequences, DNABERT \cite{DNABERT} for gene sequences, MolFormer \cite{MolFormer} for molecule SMILES strings, and BioBERT \cite{BioBERT} for descriptions of all entity types. These choices ensure that the attribute embeddings leverage domain-specific knowledge encoded in the LMs, thereby enhancing the quality and applicability of the generated embeddings. During training, the LMs are frozen to reduce the number of trainable parameters. Another concern is that Knowledge Graphs are often incomplete due to undisclosed or overlooked facts. For these cases, we randomly initialize attribute embeddings.

\subsection{Modality Fusing} \label{sec:fusion_module}

On top of proposing a collection of features collectively representing each node type, we propose a Fusion Module designed to effectively integrate diverse modalities of node-specific features into a common embedding space. Formally, for an entity \( v_i \in V \) with modality-specific embeddings \( \mathbf{x}_1, \mathbf{x}_2, \dots, \mathbf{x}_M \), where each \( \mathbf{x} \in \mathbb{R}^{d} \). The encoder function \( \mathcal{E} \) projects a concatenation of these embeddings in the space \(\mathbb{R}^{d \times M}\) into a common embedding space \(\mathbb{R}^d\), producing a unified embedding \(\mathbf{u}_i\) as follows:
%\begin{equation}
$$
\mathbf{u}_i = \mathcal{E}(\mathbf{x}_1, \mathbf{x}_2, \dots, \mathbf{x}_M), \quad \mathbf{u}_i \in \mathbb{R}^d,
$$
%\end{equation}
where each \( \mathbf{u}_i \in \mathbb{R}^D \). This approach allows each modality to be represented in the same dimensional space, facilitating further analysis or fusion at a subsequent stage of the model.

To achieve the integration of these modality-specific embeddings effectively, we utilize Attention Fusion \cite{Attention} and Relation-guided Dual Adaptive Fusion (ReDAF) \cite{ReDAF}. These fusion methods determine the contribution of each modality before combining them into a unified representation, which is essential because different modalities may carry varying levels of importance depending on the context. By assigning appropriate weights to each modality, the model can better capture the most relevant information, resulting in a more accurate and meaningful representation of the entity. Regardless of the fusion method used, a simple mean operation is applied at the final stage to ensure a balanced integration of the multi-modal embeddings, allowing for a cohesive representation of each entity.

\subsection{Graph Contrastive Learning} \label{sec:relationship_projection}

We employ GCL models to maximize the agreement between two augmented views of the same graph, facilitating the extraction of valuable insights among nodes of identical types. We specifically explore various GCL models that are suitable for Knowledge Graphs, including Deep Graph Infomax (DGI) \cite{DGI}, Graph Group Discrimination (GGD) \cite{GGD}, and Graph Contrastive Representation Learning (GRACE) \cite{GRACE}. Each of these models employs different strategies for training using the contrastive learning approach. Regarding augmentation techniques, while the diffusion method has demonstrated superior effectiveness \cite{MVGRL}, it also demands more execution time compared to alternatives. Therefore, for the sake of efficiency, we opt to mask out nodes and remove edges for quick experimentation randomly.

Formally, let the contrastive learning function \( \mathcal{C}(\cdot) \), we map a unified embedding \( \mathbf{u}_i \) to a new embedding \( \mathbf{z}_i \) with dimension \( k \). This process is defined as follows:

%\begin{equation}
$$
\mathbf{z}_i = \mathcal{C}(\mathbf{u}_i) \quad \text{where} \quad \mathbf{z}_i \in \mathbb{R}^k.
$$
%\end{equation}

% \subsubsection{Training and Optimization}

% Training involves adjusting node embeddings via backpropagation based on the contrastive loss:
% \begin{equation}
%     \text{loss} = L(z_{\text{pos}}, z_{\text{neg}})
% \end{equation}
% where \( L \) represents the chosen contrastive loss function. The model parameters are updated to minimize this loss, guided by adaptive learning rate schedules such as cosine or linear warm-up, which are critical for efficient convergence.

\subsection{Link Prediction in KG Embedding}
\label{sec:model_architecture}

KG Embedding involves an embedding function \(e:E \cup R \rightarrow \mathcal{X}\), which maps entities and relations in a knowledge graph to elements within an embedding space $\mathcal{X}$. Additionally, it includes a scoring function \(f: \mathcal{X}^3 \rightarrow \mathbb{R}\) that, given the embeddings of entities and relations in a triple, computes a score indicating the likelihood or validity of the triple. In our experiment, we utilize Relational Graph Convolutional Network (RGCN) \cite{RGCN} as the encoder to extract embeddings from graph-structured data that includes relational information. We then employ DistMult \cite{dismult} as a scoring function to map entities and relations to vector scores. The essence of the link prediction task lies in classifying the existence of edges between entities, where positive edges are drawn from the dataset and negative edges are randomly sampled. Binary Cross Entropy (BCE) loss is employed to evaluate the effectiveness of the classification as follows:
%\begin{equation}
$$
\mathcal{L}_{\text{BCE}} = -\frac{1}{N}\sum_{i=1}^{N} \left[ \mathbf{y}_{i} \log(\hat{y}) + (1 - \mathbf{y}_{i}) \log(1 - \hat{y}) \right].
$$
%\end{equation}
The regularization term is added to avoid overfitting, and is given by the sum of squared norms of the latent representations and the relation embeddings:
%\begin{equation}
$$
\mathcal{L}_{\text{reg}} = \lambda \left( \|\mathbf{X}\|^2 + \|\mathbf{Z}\|^2 \right),
$$
%\end{equation}
where $\mathbf{X}$ is the representation after being processed through the encoder and $\mathbf{Z}$ denotes relation embeddings. The final loss function is the combination of the binary cross-entropy loss and the weighted regularization term:
%\begin{equation}
$$
\mathcal{L} = \mathcal{L}_{\text{BCE}} + \alpha \mathcal{L}_{\text{reg}}.
$$
%\end{equation}
To facilitate effective batch-wise training, we utilize the GraphSAINT sampling method \cite{graphsaint} for graph sampling. This approach employs the Random Walks technique to sample subgraphs, ensuring the presence of existing edges within each batch for the link prediction task.
\section{Experiments} \label{sec:experiments}

\subsection{Experimental Setup} \paragraph{Material:} In our experiments, we utilize two principal datasets: PrimeKG++ and the DrugBank drug-target interaction dataset \cite{knox2024drugbank}. PrimeKG++ serves as our primary dataset, enriched with detailed attribute information across a variety of biological entities, making it highly suitable for comprehensive model training and evaluation. The DrugBank dataset, a curated biomedical knowledge graph, focuses specifically on drug-target protein interactions. It comprises 9,716 FDA-approved drugs and 846 protein targets, encompassing a different set of relations and nodes compared to PrimeKG++. However, the DrugBank dataset originally lacked node attributes, necessitating augmentation by incorporating detailed attribute information akin to that in PrimeKG++, thereby ensuring a comprehensive evaluation and robust performance of our model. By leveraging the enriched attribute information integrated into both datasets, we aim to thoroughly evaluate our framework's ability to handle both broad and domain-specific biomedical knowledge graphs, enabling a rigorous assessment of its performance and generalizability.

\paragraph{Comparative Analysis of Embedding Techniques on PrimeKG++:} With the introduction of PrimeKG++, our augmented dataset, we conducted a comprehensive evaluation of our approach by exploring a variety of widely-used configurations. We experimented with three well-established GCL models: Graph Gaussian Diffusion (GGD), Graph Contrastive Representation Learning (GRACE), and Deep Graph Infomax (DGI). Additionally, we examined different attribute fusion methods, including Attention Fusion and Relation-guided Dual Adaptive Fusion (ReDAF), which weigh each modality differently before fusion. As a baseline, we also included a simple fusion approach ("None") where embeddings from various modalities were combined using a mean operation without explicit weighting. To provide additional context, we compared these configurations against models trained with Random Initialization and direct Language Model (LM)-derived embeddings. Rather than focusing on identifying a single optimal configuration, our objective was to demonstrate the versatility and robustness of the proposed approach across widely-used methods. We experimented with different configurations to showcase how our framework can be applied in diverse settings. While the choice of components may depend on the specific characteristics of the dataset, our intention was to highlight the adaptability of our framework, ensuring it performs effectively under multiple configurations.

% \paragraph{Evaluating Generalizability on the DrugBank Dataset:} To further validate the robustness and generalizability of our proposed framework, we conduct comprehensive experiments on the DrugBank drug-target interaction (DTI) dataset. Similar to the evaluation on PrimeKG++, we report results for various configurations of our framework, including different GCL models and attribute fusion methods, alongside baselines like Random Initialization and direct LM-derived embeddings. In this setup, we use GCL models pretrained on PrimeKG++ to generate initial embeddings, which serve as a foundation for further fine-tuning. Subsequently, we fine-tune the Knowledge Graph Embedding (KGE) models, specifically optimized for each configuration, using the training set of the DrugBank DTI dataset. This two-step process ensures that the pretrained embeddings capture rich semantic and relational information from PrimeKG++ while adapting to the specific relational and attribute structures of DrugBank during fine-tuning. By examining the performance across different configurations, we aim to provide insights into the effectiveness of our approach for handling novel entities and demonstrate its capability to generalize across datasets with distinct relational and attribute characteristics.

\paragraph{Evaluating Generalizability on the DrugBank Dataset:} To assess the robustness and generalizability of our framework, we conducted extensive experiments on the DrugBank drug-target interaction (DTI) dataset. Our approach utilizes GCL models pretrained on PrimeKG++ to generate initial embeddings, providing a rich semantic and relational foundation. These embeddings are then fine-tuned using Knowledge Graph Embedding (KGE) models, specifically optimized for each configuration, on the training set of the DrugBank DTI dataset. This two-step process ensures that the pretrained embeddings effectively capture meaningful information from PrimeKG++ while adapting to the unique relational and attribute structures of DrugBank. By evaluating performance across various configurations, we demonstrate our framework's ability to generalize to novel entities and its effectiveness in handling datasets with diverse relational and attribute characteristics.

\paragraph{Implementation Details:} For our experiments, we randomly split the edges of PrimeKG++ and the DrugBank drug-target interaction dataset into three subsets: training, validation, and testing, with a corresponding ratio of 60:20:20. This ensures a balanced and comprehensive evaluation of our model across both datasets. The PrimeKG++ dataset provides a richly augmented set of node attributes, while the DrugBank dataset serves as a complementary benchmark for evaluating the model's generalizability to unseen nodes and distinct relational structures. In both cases, consistent hyperparameters and settings were applied to ensure a fair and rigorous evaluation process.

To further challenge the model and assess its robustness, we adjust the negative sampling ratio in our experiments. While the standard ratio is 1:1 (one negative sample for each positive sample), we increase this ratio to 1:3 and 1:5 in certain configurations. These higher ratios create significantly more difficult tasks by introducing a larger set of negative edges, testing the model's ability to distinguish true interactions from a broader range of false ones. This adjustment enables a deeper evaluation of the model's performance in scenarios closer to real-world conditions, where true interactions are relatively sparse.

The reported results are based on models with the lowest validation loss observed during training, evaluated over 100 epochs. The statistics of the dataset splits are summarized in \cref{tab:dataset_statistics}. Our model implementations are built using PyTorch and trained on a single NVIDIA A100 GPU over 3 hours for training. This setup ensures a rigorous and reproducible evaluation framework for assessing the performance and generalizability of our proposed methods.

\begin{table}[ht]
\caption{Statistics of Triple Splits for PrimeKG++ and DPI Benchmark.}
\label{tab:dataset_statistics}
\resizebox{\columnwidth}{!}{
\begin{tabular}{lcccc}
\toprule
\textbf{Dataset} & \textbf{Total} & \textbf{Training} & \textbf{Validation} & \textbf{Testing} \\
\midrule
PrimeKG++ & 3,527,861 & 2,116,717 & 705,572 & 705,572 \\
DTI benchmark & 42,012 & 25,208  & 8,402 & 8,402  \\
\bottomrule \\
\end{tabular}
}
% \small {* Note: The statistics for the DPI benchmark refer to the training and validation subsets used in each split of cross-validation.} 
\end{table}

\paragraph{Evaluation Metrics} To assess the effectiveness of our model in the link prediction task, we employ two widely recognized metrics:  Average Precision (AP) and F1-score. AP provides a comprehensive measure of precision across recall levels, making it suitable for imbalanced datasets and varying negative sampling ratios. F1-score, the harmonic mean of precision and recall, captures the balance between false positives and false negatives, offering an interpretable measure of classification performance. These metrics ensure a robust assessment of the model's effectiveness in link prediction tasks across diverse experimental settings.

\subsection{Results and Discussion}

\subsubsection{PrimeKG++}

As shown in \cref{tab:primekg_results}, embeddings derived from pre-trained Language Models (LMs) outperform those from random initialization, emphasizing the value of external knowledge in link prediction tasks. However, our proposed approach, combining contextual information from LMs with relational insights via Graph Contrastive Learning (GCL), delivers the best performance across all configurations and negative sampling ratios.

Notably, GRACE with ReDAF achieves the highest AP of 0.996 and F1 of 0.983 at a 1:1 ratio. This strong performance remains robust under challenging conditions, with AP and F1 scores of 0.988 and 0.947 at a 1:3 ratio, and 0.980 and 0.916 at a 1:5 ratio, respectively. Consistent performance across configurations is attributed to the unified link prediction model (RGCN), with variations arising from GCL setups and attribute fusion techniques.

Our framework demonstrates adaptability and robustness, excelling across multiple GCL methods (e.g., GRACE, GGD, DGI) and attribute fusion strategies. While GCL configuration differences are modest, the most significant gains appear over baseline methods. For example, LM-derived embeddings achieve an AP of 0.993 and F1 of 0.975 at a 1:1 ratio but degrade more sharply at higher ratios. In contrast, our approach consistently outperforms baselines, showcasing the effectiveness of integrating GCL with LM-derived embeddings for superior link prediction in complex biomedical knowledge graphs. 

\subsubsection{DrugBank DTI}

As shown in \cref{tab:dpi_results}, we evaluate our framework's performance on the DrugBank DTI dataset, focusing on handling unseen nodes and outperforming baselines. Embeddings derived from pre-trained Language Models (LMs) achieve strong results, with an AP of 0.994 and F1 of 0.957 at a 1:1 negative sampling ratio, highlighting the value of external knowledge.

Among the configurations, GRACE with None and Attention fusion strategies achieves the best performance, with an AP of 0.994 and F1 of 0.972 at a 1:1 ratio. These configurations remain robust at higher negative sampling ratios, with GRACE + Attention attaining an AP of 0.986 and F1 of 0.927 at 1:3 and an AP of 0.976 and F1 of 0.887 at 1:5. Other configurations, such as GGD and DGI with ReDAF, are competitive but slightly lower. For instance, GGD + ReDAF achieves an AP of 0.9865 and F1 of 0.954 at 1:1, declining to 0.965 and 0.877 at 1:3.

While some configurations do not surpass LM-derived embeddings, GRACE consistently excels, demonstrating its ability to generalize effectively to novel datasets like DrugBank. These findings underscore our framework's robustness, particularly its capacity to adapt to diverse relational and attribute structures. By leveraging both semantic and relational information, our approach consistently outperforms random initialization and remains resilient under challenging conditions.

\begin{table*}[ht]
\centering
\caption{Link prediction performance on the PrimeKG++ dataset with varying negative sampling ratios.}
\label{tab:primekg_results}
\resizebox{\textwidth}{!}{
\begin{tabular}{@{}cccccccccc@{}}
\toprule
\textbf{Initial Embedding} & \textbf{Attribute Fusion} & \textbf{GCL Models} & \multicolumn{2}{c}{\textbf{1:1}} & \multicolumn{2}{c}{\textbf{1:3}} & \multicolumn{2}{c}{\textbf{1:5}} \\ 
\cmidrule(lr){4-5} \cmidrule(lr){6-7} \cmidrule(lr){8-9}
& & & \textbf{AP} & \textbf{F1} & \textbf{AP} & \textbf{F1} & \textbf{AP} & \textbf{F1} \\ 
\midrule
Random Initialization        & -                                & -                         & 0.980 & 0.960 & 0.945 & 0.893 & 0.909 & 0.829 \\
Direct LM-derived            & None                             & -                         & 0.993 & 0.975 & 0.982 & 0.934 & 0.972 & 0.902 \\
\midrule
\multirow{9}{*}{Our Approaches}
                            & None     &                                                    & 0.993 & 0.978 & 0.979 & 0.933 & 0.966 & 0.895 \\
                            & Attention                         & GGD                       & 0.994 & 0.979 & 0.982 & 0.937 & 0.970 & 0.901 \\
                            & ReDAF                             &                           & 0.993 & 0.978 & 0.981 & 0.934 & 0.968 & 0.896 \\
\cmidrule{2-9}
                            & None                              &                           & \textbf{0.996} & \textbf{0.983} & 0.987 & \textbf{0.947} & 0.979 & 0.916 \\
                            & Attention                         & GRACE                     & \textbf{0.996} & \textbf{0.983} & 0.982 & 0.937 & \textbf{0.980} & \textbf{0.917} \\
                            & ReDAF                             &                           & \textbf{0.996} & \textbf{0.983} & \textbf{0.988} & \textbf{0.947} & 0.980 & 0.916 \\
\cmidrule{2-9}
                            & None                              &                           & 0.993 & 0.979 & 0.980 & 0.936 & 0.968 & 0.899 \\
                            & Attention                         & DGI                       & 0.994 & 0.979 & 0.982 & 0.936 & 0.970 & 0.898 \\
                            & ReDAF                             &                           & 0.993 & 0.977 & 0.979 & 0.931 & 0.965 & 0.891 \\
\bottomrule
\end{tabular}}
\end{table*}

\begin{table*}[hbt!]
    \centering
    \caption{Link prediction performance on the DrugBank DTI dataset with varying negative sampling ratios.}
    \label{tab:dpi_results}
    \resizebox{\textwidth}{!}{
        \begin{tabular}{@{}lcccccccc@{}}
            \toprule
            \textbf{Initial Embedding} & \textbf{Attribute Fusion} & \textbf{GCL Models} &  \multicolumn{2}{c}{\textbf{1:1}} & \multicolumn{2}{c}{\textbf{1:3}} & \multicolumn{2}{c}{\textbf{1:5}} \\ 
            \cmidrule(lr){4-5} \cmidrule(lr){6-7} \cmidrule(lr){8-9}
            & & & \textbf{AP} & \textbf{F1} & \textbf{AP} & \textbf{F1} & \textbf{AP} & \textbf{F1} \\
            \midrule
            Random Initialization       & -                             & -                        & 0.834 & 0.749 & 0.661 & 0.513 & 0.579 & 0.591 \\
            Direct LM-derived           & None                          & -                        & 0.994 & 0.957 & \textbf{0.988} & 0.884 & \textbf{0.982} & 0.822 \\
            \midrule
            \multirow{9}{*}{Our Approaches}
                                        & None                          &                           & 0.985 & 0.948 & 0.963 & 0.862 & 0.936 & 0.793 \\
                                        & Attention                     & GGD                       & 0.9862 & 0.951 & 0.964 & 0.870 & 0.940 & 0.803 \\
                                        & ReDAF                         &                           & 0.9865 & 0.954 & 0.965 & 0.877 & 0.941 & 0.813 \\
            \cmidrule{2-9}
                                        & None                          &                           & \textbf{0.994} & \textbf{0.972} & 0.985 & \textbf{0.928} & 0.976 & \textbf{0.887} \\
                                        & Attention                     & GRACE                     & \textbf{0.994} & \textbf{0.972} & 0.986 & 0.927 & 0.976 & \textbf{0.887} \\
                                        & ReDAF                         &                           & \textbf{0.994} & 0.969 & 0.986 & 0.918 & 0.977 & 0.871 \\
            \cmidrule{2-9}
                                        & None                          &                           & 0.986 & 0.948 & 0.964 & 0.863 & 0.940 & 0.793 \\
                                        & Attention                     & DGI                       & 0.986 & 0.95 & 0.966 & 0.870 & 0.943 & 0.803 \\
                                        & ReDAF                         &                           & 0.983 & 0.946 & 0.957 & 0.858 & 0.928 & 0.785 \\
            \bottomrule
        \end{tabular}}
\end{table*}

\subsection{Latent Space Visualization of Embeddings}

To assess embedding quality, we performed a latent space visualization using the PrimeKG++ dataset, which was used during GCL model pretraining. Visualizing the entire dataset is challenging due to the complexity of link prediction tasks and the difficulty in interpreting dense patterns. Therefore, we concentrated on the protein with the highest number of interactions, allowing us to present a focused and meaningful visualization that reflects the relational and semantic structure relevant to the link prediction objective.

Using t-SNE, we projected the high-dimensional drug embeddings into a 2D space. The embeddings were categorized into two groups: drugs interacting with the selected protein and drugs not interacting with the protein. To evaluate the effectiveness of our approach, we compared embeddings generated through two configurations: Language Model (LM)-based embeddings and embeddings enhanced through our proposed method. For our approach, we employed GRACE + ReDAF, which is our most stable configuration, effectively combining LM and Graph Contrastive Learning (GCL) to incorporate relational information.

\begin{figure*}[ht]
    \centering
    \includegraphics[width=0.7\textwidth]{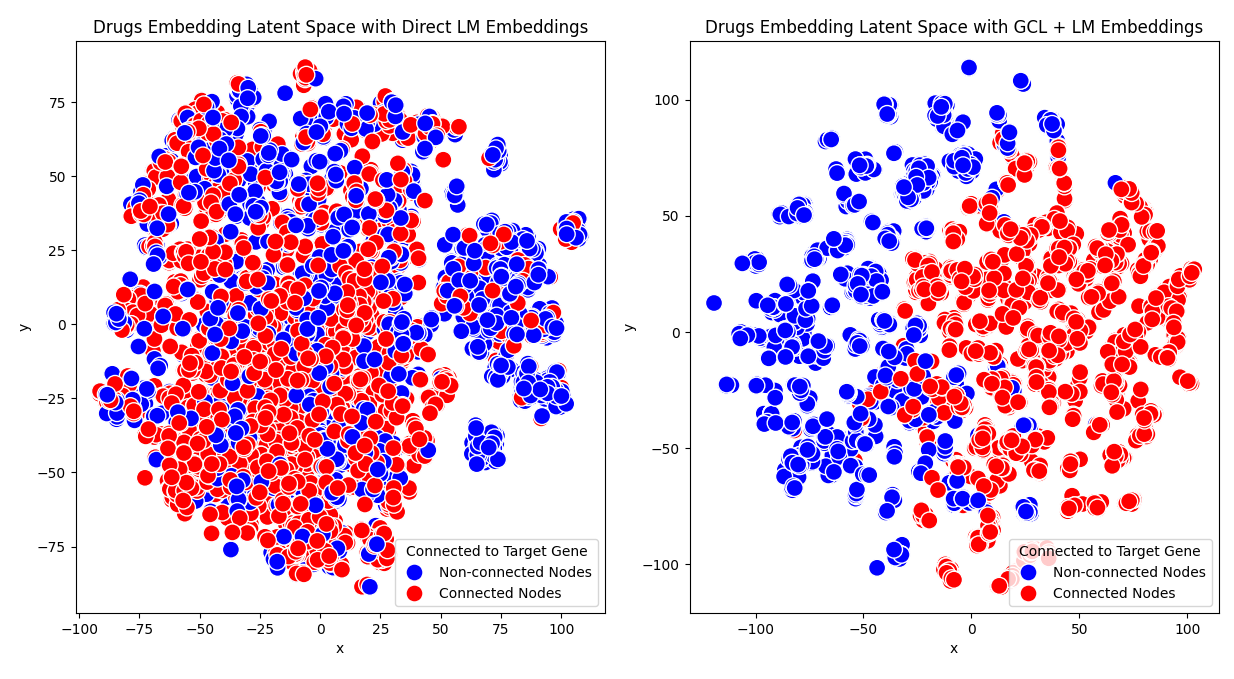}
       \caption{t-SNE visualization of drug embeddings for a single protein with the highest number of interactions in the PrimeKG++ dataset. The left panel displays embeddings derived solely from the Language Model (LM), while the right panel shows embeddings generated using our proposed approach (GRACE + ReDAF). Drugs interacting with the selected protein are labeled in red, and non-interacting drugs are labeled in blue. This comparison illustrates the structural differences in the latent space resulting from the two embedding methods.}
    \label{fig:latent_space_comparison}
    %\vspace{-0.5cm}
\end{figure*}

The visualization results in \cref{fig:latent_space_comparison} reveal notable differences between the two configurations. Embeddings generated solely with Language Models (LM) showed less distinct clustering, with considerable overlap between the two groups. This overlap suggests a limited capacity to distinguish drugs interacting with the selected protein from those that do not. In contrast, embeddings produced using our proposed method, which integrates LM with Graph Contrastive Learning (GCL), exhibited tighter clustering and more pronounced separation. This demonstrates the method's superior ability to capture shared properties among drugs interacting with the same protein. These results underscore the robustness of our framework in generating high-quality, interpretable embeddings that accurately represent underlying biological relationships, even when applied to unseen datasets such as DrugBank.

\section{Conclusion} \label{sec:conclusion}
In this paper, we presented a novel pretrained node representation model designed to enhance link prediction performance in Biomedical Knowledge Graphs (BKGs). Our approach combines semantic information from node attributes with relational data from PrimeKG++, producing robust and meaningful node embeddings. By incorporating multimodal data—such as biological sequences and textual descriptions—we enrich the contextual understanding of relationships within the graph. Furthermore, we leveraged Graph Contrastive Learning (GCL) in combination with Language Models (LMs) to optimize intra-node relationships, resulting in more generalizable embeddings capable of handling unseen nodes.

To address the issue of sparse node attributes in existing BKGs, we introduced PrimeKG++, an enriched biomedical knowledge graph that integrates biological sequences and detailed textual descriptions across various entity types. This enhancement not only resolves the limitations of PrimeKG but also serves as a valuable resource for advancing research in the field. Additionally, experiments conducted on PrimeKG++ demonstrate that our pretrained node representations significantly outperform baselines, including random initialization and direct LM-derived embeddings, highlighting the advantage of combining semantic and relational information for improved link prediction.

To further validate our framework, we evaluated it on the DrugBank drug-target interaction (DTI) dataset, showcasing its strong generalization capabilities. Despite the dataset's distinct set of relations and unseen nodes, our approach consistently outperformed baseline methods, demonstrating robust performance even under more challenging scenarios. Importantly, while this work focused on drug-protein interactions as the primary use case, the flexibility of our framework allows it to be easily extended to other relationship types, such as drug-disease or protein-disease interactions, further broadening its applicability.

This work makes substantial contributions to the field, particularly through the development of PrimeKG++, a comprehensive multimodal knowledge graph that integrates detailed biological sequences and textual descriptions, addressing key limitations of prior datasets. Our pretrained node attributes encoder, which will be made publicly available, provides a valuable tool for researchers, enabling them to directly leverage high-quality embeddings for their own work. The versatility and adaptability of our framework make it well-suited for application across diverse multimodal knowledge graphs, underscoring its broader impact in advancing biomedical knowledge representation and discovery.

\bibliography{main}

\begin{thebibliography}{10}

\bibitem{ProteinBERT}
Nadav Brandes, Dan Ofer, Yam Peleg, Nadav Rappoport, and Michal Linial.
\newblock {ProteinBERT: a universal deep-learning model of protein sequence and
  function}.
\newblock {\em Bioinformatics}, 38(8):2102--2110, 02 2022.

\bibitem{PrimeKG}
Payal Chandak, Kexin Huang, and Marinka Zitnik.
\newblock Building a knowledge graph to enable precision medicine.
\newblock {\em Scientific Data}, 10(1):67, 2023.

\bibitem{LMs_knowledge_bases5}
Huajun Chen.
\newblock Large knowledge model: Perspectives and challenges, 2023.

\bibitem{BARTSmiles}
Gayane Chilingaryan, Hovhannes Tamoyan, Ani Tevosyan, Nelly Babayan, Lusine
  Khondkaryan, Karen Hambardzumyan, Zaven Navoyan, Hrant Khachatrian, and Armen
  Aghajanyan.
\newblock Bartsmiles: Generative masked language models for molecular
  representations, 2022.

\bibitem{Dalla-Torre2023.01.11.523679}
Hugo Dalla-Torre, Liam Gonzalez, Javier Mendoza-Revilla, Nicolas~Lopez
  Carranza, Adam~Henryk Grzywaczewski, Francesco Oteri, Christian Dallago, Evan
  Trop, Bernardo~P. de~Almeida, Hassan Sirelkhatim, Guillaume Richard, Marcin
  Skwark, Karim Beguir, Marie Lopez, and Thomas Pierrot.
\newblock The nucleotide transformer: Building and evaluating robust foundation
  models for human genomics.
\newblock {\em bioRxiv}, 2023.

\bibitem{BioBLP}
Daniel Daza, Dimitrios Alivanistos, Payal Mitra, Thom Pijnenburg, Michael
  Cochez, and Paul Groth.
\newblock Bioblp: a modular framework for learning on multimodal biomedical
  knowledge graphs.
\newblock {\em Journal of Biomedical Semantics}, 14(1):20, 2023.

\bibitem{bert}
Jacob Devlin, Ming-Wei Chang, Kenton Lee, and Kristina Toutanova.
\newblock {BERT}: Pre-training of deep bidirectional transformers for language
  understanding, June 2019.

\bibitem{biolp5}
Haitao Fu, Feng Huang, Xuan Liu, Yang Qiu, and Wen Zhang.
\newblock {MVGCN: data integration through multi-view graph convolutional
  network for predicting links in biomedical bipartite networks}.
\newblock {\em Bioinformatics}, 38(2):426--434, 09 2021.

\bibitem{biolp3}
Katrin H{"a}nsel, Stephen~N. Dudgeon, Kei-Hoi Cheung, Thomas~J.S. Durant, and
  Wolfgang~L. Schulz.
\newblock From data to wisdom: Biomedical knowledge graphs for real-world data
  insights.
\newblock {\em Journal of Medical Systems}, 47(1):65, May 2023.

\bibitem{MVGRL}
Kaveh Hassani and Amir~Hosein Khasahmadi.
\newblock Contrastive multi-view representation learning on graphs, 2020.

\bibitem{LMs_knowledge_bases2}
Qiyuan He, Yizhong Wang, and Wenya Wang.
\newblock Can language models act as knowledge bases at scale?, 2024.

\bibitem{DNABERT}
Yanrong Ji, Zhihan Zhou, Han Liu, and Ramana~V Davuluri.
\newblock {DNABERT: pre-trained Bidirectional Encoder Representations from
  Transformers model for DNA-language in genome}.
\newblock {\em Bioinformatics}, 37(15):2112--2120, 02 2021.

\bibitem{LMs_knowledge_bases3}
Zhengbao Jiang, Zhiqing Sun, Weijia Shi, Pedro Rodriguez, Chunting Zhou, Graham
  Neubig, Xi~Victoria Lin, Wen tau Yih, and Srinivasan Iyer.
\newblock Instruction-tuned language models are better knowledge learners,
  2024.

\bibitem{knox2024drugbank}
Craig Knox, Mike Wilson, Christen~M Klinger, Mark Franklin, Eponine Oler, Alex
  Wilson, Allison Pon, Jordan Cox, Na~Eun Chin, Seth~A Strawbridge, et~al.
\newblock Drugbank 6.0: the drugbank knowledgebase for 2024.
\newblock {\em Nucleic acids research}, 52(D1):D1265--D1275, 2024.

\bibitem{DrugBank}
Craig Knox, Mike Wilson, Christen~M Klinger, Mark Franklin, Eponine Oler, Alex
  Wilson, Allison Pon, Jordan Cox, Na~Eun~Lucy Chin, Seth~A Strawbridge,
  Marysol Garcia-Patino, Ray Kruger, Aadhavya Sivakumaran, Selena Sanford,
  Rahil Doshi, Nitya Khetarpal, Omolola Fatokun, Daphnee Doucet, Ashley
  Zubkowski, Dorsa~Yahya Rayat, Hayley Jackson, Karxena Harford, Afia Anjum,
  Mahi Zakir, Fei Wang, Siyang Tian, Brian Lee, Jaanus Liigand, Harrison
  Peters, Ruo Qi~Rachel Wang, Tue Nguyen, Denise So, Matthew Sharp, Rodolfo
  da~Silva, Cyrella Gabriel, Joshua Scantlebury, Marissa Jasinski, David
  Ackerman, Timothy Jewison, Tanvir Sajed, Vasuk Gautam, and David~S Wishart.
\newblock {DrugBank} 6.0: The {DrugBank} knowledgebase for 2024.
\newblock {\em Nucleic Acids Res.}, 52(D1):D1265--D1275, January 2024.

\bibitem{Otter-Knowledge}
Hoang~Thanh Lam, Marco~Luca Sbodio, Marcos~Martinez Gallindo, Mykhaylo Zayats,
  Raul Fernandez-Diaz, Victor Valls, Gabriele Picco, Cesar~Berrospi Ramis, and
  Vanessa Lopez.
\newblock Otter-knowledge: benchmarks of multimodal knowledge graph
  representation learning from different sources for drug discovery.
\newblock {\em arXiv preprint arXiv:2306.12802}, 2023.

\bibitem{BioBERT}
Jinhyuk Lee, Wonjin Yoon, Sungdong Kim, Donghyeon Kim, Sunkyu Kim, Chan~Ho So,
  and Jaewoo Kang.
\newblock Biobert: a pre-trained biomedical language representation model for
  biomedical text mining.
\newblock {\em Bioinformatics}, 36(4):1234--1240, 2020.

\bibitem{BioGPT}
Patrick Lewis, Myle Ott, Jingfei Du, and Veselin Stoyanov.
\newblock Pretrained language models for biomedical and clinical tasks:
  Understanding and extending the state-of-the-art.
\newblock In Anna Rumshisky, Kirk Roberts, Steven Bethard, and Tristan Naumann,
  editors, {\em Proceedings of the 3rd Clinical Natural Language Processing
  Workshop}, pages 146--157, Online, November 2020. Association for
  Computational Linguistics.

\bibitem{ESM-2}
Zeming Lin, Halil Akin, Roshan Rao, Brian Hie, Zhongkai Zhu, Wenting Lu, Nikita
  Smetanin, Robert Verkuil, Ori Kabeli, Yaniv Shmueli, Allan dos Santos~Costa,
  Maryam Fazel-Zarandi, Tom Sercu, Salvatore Candido, and Alexander Rives.
\newblock Evolutionary-scale prediction of atomic-level protein structure with
  a language model.
\newblock {\em Science}, 379(6637):1123--1130, 2023.

\bibitem{multimodal_gcl}
Zhenxi Lin, Ziheng Zhang, Meng Wang, Yinghui Shi, Xian Wu, and Yefeng Zheng.
\newblock Multi-modal contrastive representation learning for entity alignment,
  October 2022.

\bibitem{NCBI}
Donna Maglott, Jim Ostell, Kim~D Pruitt, and Tatiana Tatusova.
\newblock Entrez gene: gene-centered information at ncbi.
\newblock {\em Nucleic acids research}, 39(suppl\_1):D52--D57, 2010.

\bibitem{biolp1}
Aditya~Krishna Menon and Charles Elkan.
\newblock Link prediction via matrix factorization.
\newblock In Dimitrios Gunopulos, Thomas Hofmann, Donato Malerba, and Michalis
  Vazirgiannis, editors, {\em Machine Learning and Knowledge Discovery in
  Databases}, pages 437--452, Berlin, Heidelberg, 2011. Springer Berlin
  Heidelberg.

\bibitem{ngo2022predicting}
Khang~Nhat Ngo, Truong~Son Hy, and Risi Kondor.
\newblock Predicting drug-drug interactions using deep generative models on
  graphs.
\newblock In {\em NeurIPS 2022 AI for Science: Progress and Promises}, 2022.

\bibitem{BKGs_applications}
David~N. Nicholson and Casey~S. Greene.
\newblock Constructing knowledge graphs and their biomedical applications.
\newblock {\em Computational and Structural Biotechnology Journal},
  18:1414--1428, 2020.

\bibitem{GMI}
Zhen Peng, Wenbing Huang, Minnan Luo, Qinghua Zheng, Yu~Rong, Tingyang Xu, and
  Junzhou Huang.
\newblock {Graph Representation Learning via Graphical Mutual Information
  Maximization}, 2020.

\bibitem{LMs_knowledge_bases1}
Fabio Petroni, Tim Rockt{\"a}schel, Patrick Lewis, Anton Bakhtin, Yuxiang Wu,
  Alexander~H Miller, and Sebastian Riedel.
\newblock Language models as knowledge bases?
\newblock {\em arXiv preprint arXiv:1909.01066}, 2019.

\bibitem{MolFormer}
Jerret Ross, Brian Belgodere, Vijil Chenthamarakshan, Inkit Padhi, Youssef
  Mroueh, and Payel Das.
\newblock {Large-scale chemical language representations capture molecular
  structure and properties}.
\newblock {\em Nature Machine Intelligence}, 4(12):1256--1264, 2022.

\bibitem{RGCN}
Michael Schlichtkrull, Thomas~N. Kipf, Peter Bloem, Rianne van~den Berg, Ivan
  Titov, and Max Welling.
\newblock Modeling relational data with graph convolutional networks, 2017.

\bibitem{Rotate}
Zhiqing Sun, Zhi-Hong Deng, Jian-Yun Nie, and Jian Tang.
\newblock Rotate: Knowledge graph embedding by relational rotation in complex
  space.
\newblock {\em arXiv preprint arXiv:1902.10197}, 2019.

\bibitem{ComplEx}
Th{\'e}o Trouillon, Johannes Welbl, Sebastian Riedel, {\'E}ric Gaussier, and
  Guillaume Bouchard.
\newblock Complex embeddings for simple link prediction.
\newblock In {\em International conference on machine learning}, pages
  2071--2080. PMLR, 2016.

\bibitem{Attention}
Ashish Vaswani, Noam Shazeer, Niki Parmar, Jakob Uszkoreit, Llion Jones,
  Aidan~N. Gomez, Lukasz Kaiser, and Illia Polosukhin.
\newblock Attention is all you need, 2023.

\bibitem{DGI}
Petar Veličković, William Fedus, William~L. Hamilton, Pietro Liò, Yoshua
  Bengio, and R~Devon Hjelm.
\newblock Deep graph infomax, 2018.

\bibitem{Biokg}
Brian Walsh, Sameh~K Mohamed, and V{\'\i}t Nov{\'a}{\v{c}}ek.
\newblock Biokg: A knowledge graph for relational learning on biological data.
\newblock In {\em Proceedings of the 29th ACM International Conference on
  Information \& Knowledge Management}, pages 3173--3180, 2020.

\bibitem{LMs_survey}
Benyou Wang, Qianqian Xie, Jiahuan Pei, Zhihong Chen, Prayag Tiwari, Zhao Li,
  and Jie Fu.
\newblock Pre-trained language models in biomedical domain: A systematic
  survey.
\newblock {\em ACM Computing Surveys}, 56(3):1--52, 2023.

\bibitem{biolp4}
Mengqi Wang, Haonan Wang, Xingyu Liu, Xinhe Ma, and Baoying Wang.
\newblock Drug-drug interaction predictions via knowledge graph and text
  embedding: Instrument validation study.
\newblock {\em JMIR Med Inform}, 9(6):e28277, Jun 24 2021.

\bibitem{dismult}
Bishan Yang, Wen tau Yih, Xiaodong He, Jianfeng Gao, and Li~Deng.
\newblock Embedding entities and relations for learning and inference in
  knowledge bases, 2015.

\bibitem{kgcl2022}
Yuhao Yang, Chao Huang, Lianghao Xia, and Chenliang Li.
\newblock Knowledge graph contrastive learning for recommendation, 2022.

\bibitem{graphsaint}
Hanqing Zeng, Hongkuan Zhou, Ajitesh Srivastava, Rajgopal Kannan, and Viktor
  Prasanna.
\newblock Graphsaint: Graph sampling based inductive learning method, 2020.

\bibitem{KEGCL}
Lihui Zhang and Ruifan Li.
\newblock {KE}-{GCL}: Knowledge enhanced graph contrastive learning for
  commonsense question answering, December 2022.

\bibitem{ReDAF}
Yichi Zhang, Zhuo Chen, Lingbing Guo, Yajing Xu, Binbin Hu, Ziqi Liu, Wen
  Zhang, and Huajun Chen.
\newblock Native: Multi-modal knowledge graph completion in the wild.
\newblock {\em Authorea Preprints}, 2024.

\bibitem{LMs_knowledge_bases4}
Jun Zhao, Zhihao Zhang, Luhui Gao, Qi~Zhang, Tao Gui, and Xuanjing Huang.
\newblock Llama beyond english: An empirical study on language capability
  transfer, 2024.

\bibitem{GGD}
Yizhen Zheng, Shirui Pan, Vincent~Cs Lee, Yu~Zheng, and Philip~S. Yu.
\newblock Rethinking and scaling up graph contrastive learning: An extremely
  efficient approach with group discrimination, 2022.

\bibitem{GRACE}
Yanqiao Zhu, Yichen Xu, Feng Yu, Qiang Liu, Shu Wu, and Liang Wang.
\newblock Deep graph contrastive representation learning, 2020.

\bibitem{biolp2}
Marinka Zitnik, Monica Agrawal, and Jure Leskovec.
\newblock {Modeling polypharmacy side effects with graph convolutional
  networks}.
\newblock {\em Bioinformatics}, 34(13):i457--i466, 06 2018.

\end{thebibliography}
\bibliographystyle{plain}

\newpage
% \newpage
\appendix
\label{sec:appendix}
\section*{Appendix}
\section{Parameters and notations}

This section provides a comprehensive overview of the hyperparameters and notations used in our study. The hyperparameters crucial for configuring and optimizing our model during training are listed in \cref{tab:hyperparameters}. These parameters include settings such as learning rate, batch size, and dropout rate, which play a significant role in the model's performance and generalization. The key notations and variables used throughout our methodology are outlined in \cref{tab:notations}, defining essential elements such as the biomedical knowledge graph structure, embedding spaces, and loss functions. These tables serve as a reference to ensure clarity and consistency in our descriptions and formulations, facilitating a better understanding of the setup and execution of our experiments. We also provide a detailed summary of the models used and constructed in this study in \cref{tab:modelparameters}

\begin{table}[ht]
\label{appendix:hyper}
\centering
\caption{List of Hyperparameters}
\label{tab:hyperparameters}
\resizebox{\columnwidth}{!}{
\begin{tabular}{lll}
\toprule
\textbf{Hyperparameter} & \textbf{Description} & \textbf{Value} \\
\midrule
Learning Rate & Step size for updates & 0.001 \\
Batch Size & Samples per update & 128 \\
Epochs & Passes through dataset & 100 \\
Embed dim & Initial embedding size & 768 \\
Hidden dim & Hidden layer size & 128 \\
Hidden layers & Number of hidden layers & 2 \\
Dropout Rate & Fraction of units to drop & 0.2 \\
Embedding Dimension & Size of embeddings & 128 \\
% Attention Heads & Number of attention heads & 8 \\
Regularization Weight ($\lambda$) & Weight for regularization & 0.01 \\
% Positive-Negative Ratio & Ratio of pos to neg samples & 1:10 \\
Random Walk Length & Length of random walks & 10 \\
Random Walk Step & Number of random walks step & 1000 \\
Optimizer & Optimization algorithm & Adam \\
Learning Rate Schedule & Schedule for learning rate & Cosine Annealing \\
Warm-up Steps & Steps for learning rate warm-up & 200 \\
Activation Function & Activation function & ReLU \\
Gradient Clipping & Max gradient norm & 1.0 \\
% Weight Decay & L2 weight decay & 0.0001 \\
Early Stopping Patience & Epochs with no improvement & 3 \\
\bottomrule
\end{tabular}
}
\end{table}

\begin{table}[ht]
\centering
\caption{List of notations}
\label{tab:notations}
\resizebox{\columnwidth}{!}{
\begin{tabular}{ll}
\toprule
\textbf{Notation} & \textbf{Definition} \\
\midrule
% $G = (V, R, E, D, d)$ & Biomedical Knowledge Graph \\
$G$ & Knowledge Graph \\
$V$ & Set of nodes: $\{v_1, v_2, \dots, v_n\}$ \\
$R$ & Set of relations: $\{r_1, r_2, \dots, r_m\}$ \\
$E$ & Set of triples $(h, r, t)$: $\{(h_1, r_1, t_1), \dots, (h_k, r_k, t_k)\}$ \\
$D$ & Dataset with attributes: $\{D_1, D_2, \dots, D_p\}$ \\
$d : V_d \rightarrow D$ & Function mapping entities to attributes: $d(v_i) \rightarrow D_i$ \\
$e_i$ & Modality-specific encoder: $e_i: D_i \rightarrow \mathcal{X}_i$ \\
$\mathcal{X}_i$ & Embedding space of modality $i$: $\mathbb{R}^{d_i}$ \\
$\mathbf{x}_i$ & Modality-specific embedding: $\mathbf{x}_i \in \mathbb{R}^{d_i}$ \\
$\mathcal{E}$ & Encoder function: $\mathcal{E}(\mathbf{x}_i) \rightarrow \mathbb{R}^D$ \\
$\mathbf{h}_i$ & Unified embedding: $\mathbf{h}_i \in \mathbb{R}^D$ \\
$\mathbf{A}$ & Adjacency matrix: $\mathbf{A} \in \mathbb{R}^{n \times n}$ \\
$\mathbf{X}$ & Node feature matrix: $\mathbf{X} \in \mathbb{R}^{n \times d}$ \\
$\mathbf{Z}$ & Latent node representation: $\mathbf{Z} \in \mathbb{R}^{n \times d}$ \\
$\mathbf{R}$ & Set of relations: $\mathbf{R} \in \mathbb{R}^{m \times d}$ \\
% $\mathcal{L}_{\text{BCE}}$ & Binary cross-entropy loss: $- \frac{1}{N} \sum_{i=1}^{N} [y_i \log(\hat{y}_i) + (1 - y_i) \log(1 - \hat{y}_i)]$ \\
% $\mathcal{L}_{\text{reg}}$ & Regularization term: $\lambda (\|\mathbf{Z}\|^2 + \|\mathbf{R}\|^2)$ \\
% $\mathcal{L}$ & Total loss function: $\mathcal{L}_{\text{BCE}} + \alpha \mathcal{L}_{\text{reg}}$ \\
% $\hat{y}$ & Prediction vector: $[\hat{y}_{ij}^{+}, \hat{y}_{i'j'}^{-}]$ \\
% $y$ & Ground truth labels: $[1, 0]$ \\
$\alpha$ & Weight for regularization term: 0.01 \\
% $h$ & Head entity in a triple: $h \in V$ \\
$h$ & Head entity in a triple: $h \in V$\\
$t$ & Tail entity in a triple: $t \in V$ \\
$r$ & Relation in a triple: $r \in R$ \\
\bottomrule
\end{tabular}
}
\end{table}

\begin{table}[ht]
\centering
\caption{Model summary}
\label{tab:modelparameters}
\resizebox{\columnwidth}{!}{
\begin{tabular}{lll}
\toprule
\textbf{Category} & \textbf{Model} & \textbf{No. Parameter} \\
% \midrule

% \multirow{4}{*}{Modality encoding} & ProtBERT & 16M \\ 
% & DNABert-2 & 117M \\
% & Molformer & 313M \\ 
% & BioBERT & 110M \\
\cmidrule{1-3}
GCL & GCN Encoder & 164K \\
\cmidrule{1-3}
\multirow{2}{*}{Modality Fusion} & Attention & 1.8M \\
& ReDAF & 1.2M \\
\cmidrule{1-3}
KGE & RGCN & 590K \\
\bottomrule
\end{tabular}
}
\end{table}

\section{PrimeKG++ Creation}
\label{appendix:primekg}

To obtain and integrate the metadata, in addition to PrimeKG \cite{PrimeKG}, we perform the following steps:
\paragraph{Data Collection:}We utilize the open-source web-scraping library Selenium to fetch detailed information from authoritative sources such as Entrez Gene \cite{NCBI} and DrugBank \cite{DrugBank} for each node type. This process involves the following steps:
For genes/proteins, we collect both protein-coding genes, annotated with amino acid sequences, and non-coding genes, represented with nucleotide sequences. This information is fetched using the Entrez Gene API, which allows us to gather comprehensive gene data efficiently. For drugs, we source data from DrugBank, including molecular data represented as SMILES strings and antibody data identified by amino acid sequences. This comprehensive data collection ensures that we cover various aspects of drug representation.
Each of these new data points is categorized into sub-types, collectively representing a node type. This detailed categorization enhances the granularity and richness of our knowledge graph.

\paragraph{Data Processing:}Once the data is collected, we process it to create embeddings for each data point within the sub-types using corresponding language models (LMs). This process involves several key steps:
Embeddings for protein-coding genes and non-coding genes are generated using models trained on biological sequence data, ensuring that the embeddings capture the essential features of the sequences.
For drug data, embeddings are created using chemical language models that understand molecular structures and properties.
These embeddings aim to map each subtype into its respective embedding space. Using our proposed feature-fusion layer, we integrate sub-type embeddings of a node into a unified embedding space, resulting in a comprehensive node embedding. This fusion layer is designed to effectively combine information from different sub-types, providing a holistic representation of each node.
These unified embeddings serve as the initial embeddings for each node, capturing the rich contextual information from the various sub-types.

\paragraph{Data Storage and Usage:}All embeddings are stored in a structured format, mapping each node's name to its respective embedding. This process involves:
Organizing the embedding files in a dedicated folder, ensuring easy access and management. Each file contains embeddings for a specific sub-type, labeled accordingly.
During the training phase, we use a custom data module that loads a table storing node interactions. This module integrates the embedding tables to create a PyTorch Geometric (PyG) Heterogeneous Data object. The data module performs the following functions:
It reads the node interaction table, mapping interactions to the corresponding node embeddings.
It constructs the complete graph structure, ensuring that each node is accurately mapped to its respective embedding. This involves validating the consistency and coherence of the graph structure and addressing any discrepancies in the data.
It prepares the data object for training, ensuring that all necessary attributes and relationships are correctly represented.
This data object represents the complete graph structure, with each node accurately mapped to its respective embedding, ensuring a robust foundation for training our models.

These detailed steps ensure that our enhanced knowledge graph, PrimeKG++, provides a richer and more comprehensive dataset, supporting advanced biomedical analyses.

\section{Attribute Embedding Fusion}
\label{appendix:AEF}
\subsection{Attention Fusion}

The Attention Fusion layer integrates diverse modality-specific embeddings into a unified representation by employing attention mechanisms. This approach enables the model to dynamically weigh the importance of each modality based on its relevance to the task, thus enhancing the overall quality of the integrated embeddings.

Formally, consider an entity \( v \in V \) with modality-specific embeddings \( \mathbf{x}_1, \mathbf{x}_2, \dots, \mathbf{x}_M \), where each \( \mathbf{x}_i \in \mathbb{R}^{d_i} \). The Attention Fusion layer projects these embeddings into a common space \( \mathbb{R}^D \), and then uses attention scores to combine them.

First, each modality-specific embedding \( \mathbf{x}_i \) is transformed into a common embedding space \( \mathbb{R}^D \) using a learnable projection matrix \( \mathbf{W}_i \in \mathbb{R}^{D \times d_i} \):
%\begin{equation}
$$
\mathbf{h}_i = \mathbf{W}_i \mathbf{x}_i \quad \text{for each} \quad i = 1, 2, \dots, M,
$$
%\end{equation}
where \( \mathbf{h}_i \in \mathbb{R}^D \) represents the projected embeddings.

Next, an attention mechanism computes attention scores for each projected embedding \( \mathbf{h}_i \). The attention score \( \alpha_i \) for the \( i \)-th embedding is calculated as follows:
%\begin{equation}
$$
\alpha_i = \frac{\exp(\mathbf{q}^\top \mathbf{h}_i)}{\sum_{j=1}^{M} \exp(\mathbf{q}^\top \mathbf{h}_j)},
$$
%\end{equation}
where \( \mathbf{q} \in \mathbb{R}^D \) is a learnable query vector, and \( \alpha_i \) represents the normalized attention score for the \( i \)-th embedding.

The final unified embedding \( \mathbf{h} \) is obtained by computing a weighted sum of the projected embeddings \( \mathbf{h}_i \) based on their attention scores:
%\begin{equation}
$$
\mathbf{h} = \sum_{i=1}^{M} \alpha_i \mathbf{h}_i,
$$
%\end{equation}
where \( \mathbf{h} \in \mathbb{R}^D \) is the fused representation that integrates information from all modalities.

The Attention Fusion layer effectively combines modality-specific embeddings, allowing the model to focus on the most relevant features from each modality. This process enhances the overall quality of the embeddings, making them more suitable for downstream tasks such as link prediction.

The entire process can be summarized in the following steps.

\subsection{Relation-guided Dual Adaptive Fusion (ReDAF)}

Given the sparse nature of PrimeKG++, we utilize the Relation-guided Dual Adaptive \cite{ReDAF} Fusion model which produces a joint embedding projected from weighted parameters collected from individual modal training data. Additionally, the missing values of any element are consolidated with a random vector within the same vector space.
%\begin{equation}
$$
\omega_m(v, r) = \frac{\exp(V \odot \tanh(v_m) / \sigma(\zeta_r))}{\sum_{n \in M \cup \{S\}} \exp(V \odot \tanh(v_n) / \sigma(\zeta_r))},
$$
%\end{equation}
where V is a learnable vector and $\odot$ is the point-wise operator. Tanh() is the tanh function. $\sigma$ represents the sigmoid function to limit the relational-wise temperature in (0, 1), aiming to amplify the differences between different modal weights. With the adaptive weights, the joint embedding of an entity v is aggregated as:
%\begin{equation}
$$
v_{\text{joint}} = \sum_{m \in M \cup \{S\}} \omega_m(v, r) v_m, 
$$
%\end{equation}
where \(\sigma(x) = \frac{1}{1 + e^{-x}}\) represents the sigmoid function, \(X_i\) are the input features from the \(i\)-th modality, \(S_i\) are subtype embeddings, \(W_t\) and \(W_r\) are transformation matrices for input features and relational context \(C\), respectively, and \(w_i\) are the adaptive weights for each modality. The ReLU function is fully expressed as the maximum between zero and its input, integrating the features under a non-linear transformation.

\section{Contrastive learning settings}
\label{appendix:CL}
\subsection{Deep Graph Infomax model}

% For appendix
Deep Graph Infomax (DGI) \cite{DGI} utilizes an unsupervised learning strategy for graph data by maximizing the mutual information between node representations and a global summary of the graph. The method begins with an assumption of a set of node features $X = \{\mathbf{x}_1, \mathbf{x}_2, \ldots, \mathbf{x}_N\}$, where $N$ is the number of nodes, and each $\mathbf{x}_i \in \mathbb{R}^F$ denotes the features of node $i$. These are complemented by an adjacency matrix $A \in \mathbb{R}^{N \times N}$, which encodes the relational structure between nodes.

The core of DGI is an encoder function $E: \mathbb{R}^{N \times N} \times \mathbb{R}^{N \times F} \rightarrow \mathbb{R}^{N \times F_0}$ that transforms the node features and the adjacency matrix into high-level node embeddings $\{\mathbf{h}_1, \mathbf{h}_2, \ldots, \mathbf{h}_N\}$. These embeddings, or patch representations, are meant to encapsulate not only the properties of individual nodes but also their neighborhood structures.

To capture the global structure of the graph, DGI uses a readout function $R: \mathbb{R}^{N \times F} \rightarrow \mathbb{R}^F$ to aggregate these patch representations into a summary vector $\mathbf{s} = R(E(X, A))$. This vector serves as a comprehensive representation of the entire graph's topology and feature distribution.

DGI employs a discriminator $D: s^Ts $, which evaluates the mutual information between local patch representations and the global summary by assigning probability scores. These scores indicate how well the local patches (node embeddings) and the global summary correspond to each other in terms of information content.

For training, negative samples are generated through a stochastic corruption function $C: \mathbb{R}^{N \times N} \times \mathbb{R}^{N \times F} \rightarrow \mathbb{R}^{M \times M} \times \mathbb{R}^{M \times F}$, creating perturbed versions of the graph $(X_e, A_e) = C(X, A)$. The learning objective is to discriminate between the "true" patch-summary pairs and those generated from corrupted inputs using a noise-contrastive estimation with a binary cross-entropy loss.

This setup ensures that the encoder and discriminator learn to retain and emphasize features that are important across the graph, facilitating the discovery of intricate patterns and structural roles within the network, which can significantly enhance performance on downstream tasks like node classification.

\subsection{Graph Group Discrimination model}

We experiment with a Group-discrimination-based method called Graph Group Discrimination (GGD)  \cite{GGD}. Contrastive learning in this method is formulated to discriminate between groups of node embeddings, rather than individual pairs. This method leverages a binary cross-entropy loss to effectively distinguish between node samples from `positive' (unaltered) and `negative' (altered) graph structures.

Formally, in the GGD module, a graph autoencoder framework is employed to learn embeddings that are predictive of the graph structure. Nodes $v_i \in V$ are mapped to embeddings $\mathbf{z}_i$ using a GCN encoder $\mathcal{E}$. The model then predicts the presence or absence of edges between node pairs by computing logits $\hat{y}_{ij} = \mathbf{z}_i^T \mathbf{z}_j$. The binary cross-entropy loss is used to train the model:

\begin{align}
\mathcal{L}(\theta) = & -\sum_{(v_i, v_j) \in E} \log(\sigma(\hat{y}_{ij})) \nonumber \\
                      & -\sum_{(v_i, v_j) \notin E} \log(1 - \sigma(\hat{y}_{ij})),
\end{align}
where $\sigma$ denotes the sigmoid function.

The primary advantage of GD is its efficiency, especially in scenarios involving large-scale graph datasets, where it reduces the computational overhead and accelerates the training process significantly. By applying this approach, our model can achieve rapid convergence and robust performance even with minimal training epochs.

\subsection{Graph Contrastive representation learning model}

GRACE (Graph Contrastive Representation Learning) \cite{GRACE} applies stochastic augmentations to both node features and graph structure to learn robust node embeddings. For a graph with feature matrix $\mathbf{X}$ and adjacency matrix $\mathbf{A}$, two corrupted views $\mathbf{X}_1, \mathbf{A}_1$ and $\mathbf{X}_2, \mathbf{A}_2$ are generated by independently dropping features and edges. Node embeddings for these views are computed as $\mathbf{Z}_1 = \mathcal{E}(\mathbf{X}_1, \mathbf{A}_1)$ and $\mathbf{Z}_2 = \mathcal{E}(\mathbf{X}_2, \mathbf{A}_2)$, using the same encoder $\mathcal{E}$. Each embedding vector is then projected through a two-layer network with RELU activations to align the representations from different views while maintaining discriminative features. The contrastive loss, specifically the InfoNCE loss, is applied to align these representations while also distinguishing them from negatives within their minibatch:
%\begin{equation}
$$
\mathcal{L}(\theta) = -\sum_{i=1}^n \log \frac{\exp(\mathcal{P}(\mathbf{z}_{1i})^T \mathcal{P}(\mathbf{z}_{2i}) / \tau)}{\sum_{j=1}^n \exp(\mathcal{P}(\mathbf{z}_{1i})^T \mathcal{P}(\mathbf{z}_{2j}) / \tau)},
$$
%\end{equation}
where $\tau$ is a temperature scaling parameter.

\paragraph{Loss Function}
The contrastive loss function in GRACE is designed to maximize the agreement between node embeddings across these views while minimizing agreement with other nodes' embeddings. The loss is formulated as:
%\begin{equation}
$$
\mathcal{L} = \frac{1}{2N} \sum_{i=1}^N \left[ \ell(u_i, v_i) + \ell(v_i, u_i) \right],
$$
%\end{equation}
where $u_i$ and $v_i$ are embeddings of node $i$ from two views, respectively. The pairwise loss $\ell(u, v)$ encourages similarity of embeddings from the same node and dissimilarity from others, calculated using:
%\begin{equation}
$$
\ell(u, v) = -\log \frac{\exp(\theta(u, v) / \tau)}{\sum_{k=1}^N \exp(\theta(u, v_k) / \tau)},
$$
%\end{equation}
with $\theta(u, v)$ representing the cosine similarity and $\tau$ a temperature scaling parameter.

This method aligns with the principles of mutual information maximization and triplet loss, enhancing learning efficiency and representation quality (Zhu et al., 2020).

Following the GCL framework, node embeddings are processed through a Graph Convolutional Network (GCN) to preserve topological fidelity and feature correlation. The embeddings are then subjected to a binary classification scheme where the contrastive loss is calculated, streamlining the training phase and focusing on global structure discrimination rather than detailed pairwise node comparison:

The contrastive learning process is driven by the following loss functions:
\begin{itemize}
    \item \textbf{Jensen-Shannon Divergence (JSD):} This is quantified as:
    %\begin{equation}
    $$
    L_{\text{JSD}}(P, Q) = \frac{1}{2} D(P \| M) + \frac{1}{2} D(Q \| M),
    $$
    %\end{equation}
    where \( P \) and \( Q \) are the probability distributions of the positive and negative samples, respectively, and \( M = \frac{1}{2}(P + Q) \).

    \item \textbf{Information Noise-Contrastive Estimation (InfoNCE):} 
    %\begin{equation}
    $$
    L_{\text{InfoNCE}} = -\log \frac{\exp(z \cdot z_{\text{pos}} / \tau)}{\sum_{\text{neg}} \exp(z \cdot z_{\text{neg}} / \tau)},
    $$
    %\end{equation}
    where \( z_{\text{pos}} \) and \( z_{\text{neg}} \) are embeddings of positive and negative examples, respectively, and \( \tau \) is a temperature parameter.

    \item \textbf{Binary Cross-Entropy (BCE):}
    %\begin{equation}
    %\small 
    $$
    L_{\text{BCE}} = -\left[y \log(\sigma(z)) + (1 - y) \log(1 - \sigma(z))\right],
    $$
    %\end{equation}
    where \( \sigma \) is the sigmoid function, \( z \) is the logit, and \( y \) is the label indicating positive or negative pairs.
\end{itemize}

\section{Additional Results}
\label{sec:appendix-results}
In this section, we present additional experimental results to assess the effectiveness of our proposed approach and its components. First, we examine the impact of embedding size, showing that larger embeddings lead to improved performance. Next, we evaluate precision across different relation types, demonstrating that our model performs well in distinguishing between true and false relationships. Finally, we assess embedding quality in downstream tasks, where our approach, combining intra- and inter-learning, yields better embeddings that contribute to stronger task performance. These findings offer valuable insights to support future research in this area.

\subsection{Impact of Embedding Size on Model Performance}

The size of the embedding plays a critical role in model performance, as it determines the capacity to capture complex features of the data. To identify the optimal configuration and understand the trade-off between embedding size and performance, we systematically evaluate the impact of various embedding sizes using the Grace-Attention model. This experimentation provides insights into how embedding dimensionality influences the model's capacity and effectiveness.

The results, summarized in \cref{tab:embedding_size_results}, indicate that as the embedding size increases, both F1-score and AP improve, indicating that larger embeddings capture more information, leading to better performance. However, the performance improvement between 128 and 256 is marginal, suggesting diminishing returns for increasing embedding size beyond a certain threshold. 

% Additionally, it is important to consider the computational cost associated with larger embeddings, as they require more memory and processing power, which may not be practical for all use cases.

\begin{table}[ht]
    \centering
    \caption{Impact of embedding size on link prediction performance.}
    \label{tab:embedding_size_results}
    \begin{tabular}{cccc}
    \toprule
    \textbf{\makecell{Embedding \\ Size}} & \textbf{\makecell{No. \\ Parameters}} & \textbf{AP} & \textbf{F1} \\ 
    \midrule
    64  &  258K & 0.988  & 0.97  \\
    128 &  738K & 0.994  & 0.98  \\
    256 &  2M & 0.996  & 0.983  \\
    \bottomrule
    \end{tabular}
\end{table}

\subsection{Performance Per Relation Type}
\label{sec:appendix-edge-wise-precision}
To understand how our approach generalizes across different biomedical relationships, we evaluate the model's performance for each relation type within PrimeKG++ using the Grace-Attention model. The primary evaluation metric used are Average Precision and F1-score, as it provides a stable and clear measure of performance, particularly given the variability in the number of negative edges due to random negative sampling. This allows us to assess how well our model differentiates true relationships (true positives) from incorrect predictions (false positives) across diverse types of relations. To further examine robustness and generalizability, we train and test the model using a 1:10 negative sampling ratio.

The results, summarized in \cref{tab:relation_type_precision}, present the precision values for each relation type in PrimeKG++. Our findings indicate that the Grace-Attention model maintains high precision across all relation types, regardless of the size of the relation set. Notably, the high precision in predicting drug-protein interactions suggests that the model is highly effective in identifying accurate associations between drugs and proteins, which is critical for drug repurposing. Such precise predictions can help in discovering new therapeutic uses for existing drugs and identifying potential drug interactions, ultimately supporting more targeted and efficient drug development efforts.

\begin{table}[ht]
    \centering
    \caption{Precision per relation type in PrimeKG++ using the Grace-Attention model.}
    \label{tab:relation_type_precision}
    \begin{tabular}{@{}ccc@{}}
    \toprule
    \textbf{Relation Type} & \textbf{Precision} & \textbf{\makecell{Number of \\ Positive Edges}} \\ 
    \midrule
    Contraindication   & 0.991 & 61,350 \\
    Disease-Disease    & 0.954 & 64,388 \\
    Disease-Protein    & 0.867 & 160,822 \\
    Drug-Drug          & 0.974 & 2,672,628 \\
    Drug-Protein       & 0.995 & 51,306 \\
    Indication         & 0.989 & 18,776 \\
    Off-Label Use      & 0.994 & 4,429,078 \\
    Protein-Protein    & 0.911 & 642,150 \\
    \bottomrule
    \end{tabular}
\end{table}

\subsection{Evaluating Embedding Quality for Downstream Tasks}
\label{sec:appendix-embedidng-quality}
To further assess the effectiveness of the node embeddings on downstream tasks, specifically DrugBank DTI, we initialize the embeddings with outputs from a Knowledge Graph Embedding (KGE) model and train a machine learning model using XGBoost. The XGBoost model is configured with 500 estimators and a learning rate of 0.01. To ensure robust evaluation, we use a stratified 5-fold cross-validation approach, where metrics are reported as the mean performance across all folds.  

While link prediction has been previously conducted in our study to evaluate embedding quality, it primarily focuses on reconstructing known relationships within the graph. By contrast, training a machine learning model for a downstream task allows us to assess whether the embeddings effectively capture task-specific patterns and generalize beyond the original graph structure, providing a more comprehensive evaluation of embedding quality. 

\begin{table}[ht]
    \centering
    \caption{Comparison of embedding methods for ML downstream task}
    \label{tab:embedding-performance}
    \resizebox{\columnwidth}{!}{
    \begin{tabular}{ccccc}
    \toprule
    \textbf{Embedding} & \textbf{GCL Models} & \textbf{Fusion} & \textbf{AP} & \textbf{F1} \\
    \midrule
    Random From Scratch & - & - & 0.233 & 0 \\
    Random & - & - & 0.508 & 0.509 \\
    LM & - & - & 0.555 & 0.560 \\
    \midrule
    % \multirow{9}{*}{Our approach} 
    %     & - & None & 0.612 & 0.624 \\
    %     & GGD & Attention & \textbf{0.646} & \textbf{0.656} \\
    %     & GGD & ReDAF & 0.634 & 0.651 \\
    %     \cmidrule(lr){2-5}
    %     & - & None & 0.621 & 0.601 \\
    %     & GRACE & Attention & 0.636 & 0.611 \\
    %     & GRACE & ReDAF & 0.612 & 0.608 \\
    %     \cmidrule(lr){2-5}
    %     & - & None & 0.633 & 0.625 \\
    %     & DGI & Attention & 0.640 & 0.645 \\
    %     & DGI & ReDAF & 0.639 & 0.642 \\
    \bottomrule
    \end{tabular}
    }
\end{table}

The results in \cref{tab:embedding-performance} highlight that our framework, which integrates self-supervised intra-learning through Graph Contrastive Learning (GCL) and inter-learning via the link prediction task, significantly outperforms both random initialization and Direct-LM embeddings. GCL consistently achieves higher performance, showcasing its effectiveness in capturing richer and more comprehensive embeddings.  

Compared to training from scratch, where each node is initialized randomly, our approach delivers superior results across multiple configurations, emphasizing the critical role of embedding quality in downstream tasks. For future development, this framework can serve as a baseline for initializing embeddings in machine learning models, significantly reducing resource usage while maintaining strong performance.

\end{document}